\documentclass[11pt,letterpaper]{article}
\usepackage[letterpaper]{geometry}
\usepackage{acl2012}
\usepackage{natbib}
\renewcommand{\cite}[2][]{\citep[#1]{#2}}   
 
\usepackage{times}
\usepackage{latexsym}
\usepackage{caption}
\usepackage{subcaption}
\usepackage{amsmath}
\usepackage{multirow}
\usepackage{booktabs}
\usepackage{forest}
\forestset{
    nice empty nodes/.style={
        for tree={
            s sep=0.1em, 
            l sep=0.33em,
            inner ysep=0.4em, 
            inner xsep=0.05em,
            l=0,
            calign=midpoint,
            fit=tight,
            where n children=0{
               tier=word,
               minimum height=1.25em,
            }{},
            where n children=2{
               l-=1em,
            }{},
            parent anchor=south,
            child anchor=north,
            delay={if content={}{
                    inner sep=0pt,
                    edge path={\noexpand\path [\forestoption{edge}] 
                    			(!u.parent anchor) 
                               -- (.south)\forestoption{edge label};}
                }{}}
        },
    },
}

\usepackage{url}
\usepackage{scalefnt}
\usepackage{graphicx}
\usepackage{tikz}
\usepackage{todonotes}
\usepackage{xcolor}

\makeatletter
\newcommand{\@BIBLABEL}{\@emptybiblabel}
\newcommand{\@emptybiblabel}[1]{}
\makeatother
\usepackage[hidelinks,breaklinks=true]{hyperref} 

\title{Do latent tree learning models identify meaningful structure in sentences?} 

\author{
Adina Williams$^{1}$\\
\texttt{\small adinawilliams@nyu.edu}
\And
Andrew Drozdov$^{2,}$\thanks{~~Now at eBay, Inc.} \\ 
\texttt{\small andrew.drozdov@nyu.edu}
\And
Samuel R.~Bowman$^{1,2,3}$\\
\texttt{\small bowman@nyu.edu}
\AND
$^{1}$\normalfont Dept. of Linguistics\\New York University\\10 Washington Place\\New York, NY 10003\And
$^{2}$\normalfont Dept. of Computer Science\\New York University\\60 Fifth Avenue\\New York, NY 10011\And
$^{3}$\normalfont Center for Data Science\\New York University\\60 Fifth Avenue\\New York, NY 10011
} 

\date{}

\begin{document}
\maketitle
\begin{abstract}
Recent work on the problem of \textit{latent tree learning} has made it possible to train neural networks that learn to both parse a sentence and use the resulting parse to interpret the sentence, all without exposure to ground-truth parse trees at training time.
Surprisingly, these models often perform better at sentence understanding tasks than models that use parse trees from conventional parsers. This paper aims to investigate what these latent tree learning models learn. We replicate two such models in a shared codebase and find that (i) only one of these models outperforms conventional tree-structured models on sentence classification, (ii) its parsing strategies are not especially consistent across random restarts, (iii) the parses it produces tend to be shallower than standard Penn Treebank (PTB) parses, and (iv) they do not resemble those of PTB or any other semantic or syntactic formalism that the authors are aware of. 
\end{abstract}

\section{Introduction}        

Tree-structured recursive neural networks \citep[TreeRNNs;][]{socher2011semi}---which build a vector representation for a sentence by incrementally computing representations for each node in its parse tree---have been proven to be effective at sentence understanding tasks like sentiment analysis \citep{socher-EtAl:2013:EMNLP}, textual entailment \citep{bowman2016spinn}, and translation \citep{eriguchi-hashimoto-tsuruoka:2016:P16-1}. Some variants of these models \citep{socher2011semi,bowman2016spinn} can also be trained to produce  parse trees that they then consume. Recent work on \textit{latent tree learning} \citep{yogatama2016learning,maillard2017jointly,choi2017unsupervised} has led to the development of new training methods for TreeRNNs that allow them to learn to parse without ever being given an example of a correct parse tree, thereby replacing direct syntactic supervision with indirect supervision from a downstream task like sentence classification. These models are designed to learn grammars---strategies for assigning trees to sentences---that are suited to help solve the sentence understanding task at hand, rather than ones that approximate expert-designed grammars like that of the Penn Treebank \citep[PTB;][]{ptb}.
 
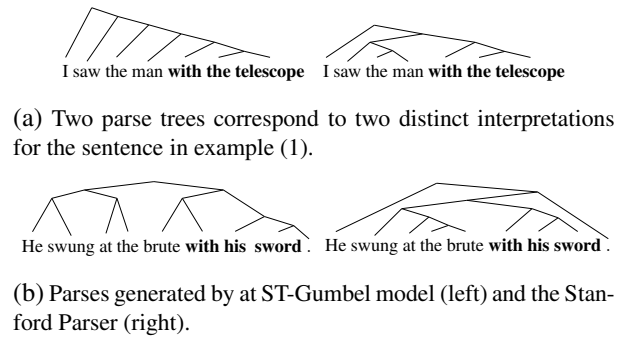
\begin{figure}[t]
\centering

\begin{subfigure}[t]{\columnwidth}\centering

\scalebox{0.63}{
\begin{forest}
 shape=coordinate,
 where n children=0{
   tier=word
 }{},
 nice empty nodes
[[I] [ [saw] [[the] [ [man] [[\textbf{with}] [[\textbf{the}][\textbf{telescope}]]]]]]]
\end{forest}
}
\scalebox{0.63}{
\begin{forest}
 shape=coordinate,
 where n children=0{
   tier=word
 }{},
 nice empty nodes
[[I] [ [ [saw] [[the] [man] ] ] [[\textbf{with}] [[\textbf{the}][\textbf{telescope}]]]]]
\end{forest}
}
\caption{\footnotesize{Two parse trees correspond to two distinct interpretations for the sentence in example (\ref{tele}).}}
\end{subfigure}\\\vspace{0.5em}
\begin{subfigure}[t]{\columnwidth}\centering
\scalebox{0.63}{
\begin{forest}
 shape=coordinate,
 where n children=0{
   tier=word
 }{},
 nice empty nodes
[ [ [ [He] [swung] ] [ [at] [the] ] ] [ [ [brute] [\textbf{with}] ] [ [\textbf{his}] [, phantom ] [ [\textbf{sword}] [$.$] ] ] ] ]
\end{forest}}
\scalebox{0.63}{
\begin{forest}
 shape=coordinate,
 where n children=0{
   tier=word
 }{},
 nice empty nodes
[ [He] [ [ [ [swung] [ [at] [ [the] [brute] ] ] ] [ [\textbf{with}] [ [\textbf{his}] [\textbf{sword}] ] ] ] [$.$] ] ]
\end{forest}}%
\caption{\footnotesize{Parses generated by at ST-Gumbel model (left) and the Stanford Parser (right).}}
\end{subfigure}
 \caption{\label{fig:exampletrees} Examples of unlabeled binary parse trees.} 
\end{figure}

Latent tree learning models have shown striking success at sentence understanding, reliably performing better on sentiment analysis and textual entailment than do comparable TreeRNN models which use parses assigned by conventional parsers, and setting the state of the art among sentence-encoding models for textual entailment. However, none of the work in latent tree learning to date has included any substantial evaluation of the quality of the trees induced by these models, leaving open an important question which this paper aims to answer: Do these models owe their success to consistent, principled latent grammars?  If they do, these grammars may be worthy objects of study for researchers in syntax and semantics. If they do not, understanding why the models succeed without such syntax could lead to new insights into the use of TreeRNNs and into sentence understanding more broadly.

While there is still lively debate within linguistic syntax and semantics over the precise grammars that should be used for language understanding and generation, it has long been clear that understanding any natural language sentence requires implicitly or explicitly recognizing which substrings of the sentence form meaningful units, or \textit{constituents} \citep{chomskyaspects,fregesinn,kratzer1998}. This is well illustrated by structurally ambiguous sentences like the one below, repeated from \citet{sag1991} a.o.:
\begin{enumerate}
    \item \label{tele}
    \begin{enumerate}
        \setlength\itemsep{-.25em}
         \item {I saw the [ man [ with the telescope ] ]} \\ $\hookrightarrow$I saw the man who had a telescope.\label{tele2}
        \item {I [ saw the man ] [ with the telescope ]} \\ $\hookrightarrow$ I used the telescope to view the man.\label{tele1}
     \end{enumerate}
\end{enumerate}

Under the partial constituency parse shown in \ref{tele2}, \textit{with a telescope} forms a constituent with \textit{man}, thereby providing additional information about the individual \textit{man} describes. On the other hand, in \ref{tele1}, \textit{with a telescope} does not form a constituent with \textit{man}, but instead provides additional information about the action described by \textit{saw a man}. In this way, the same string of words can be assigned two different, yet equally valid constituency structures reflecting the different interpretations for the string. Constituency can be straightforwardly expressed using unlabeled parse trees like the ones used in TreeRNNs, and expressing constituency information is generally thought to be the primary motivation for using trees in TreeRNNs. 

In this paper, we reimplement the latent tree learning models of \citet{yogatama2016learning} and \citet{choi2017unsupervised} in a shared codebase, train both models (and several baselines) to perform textual entailment on the SNLI and MultiNLI corpora \citep{bowman-EtAl:2015:EMNLP,williams2017broad}, and evaluate the results quantitatively and qualitatively with a focus on four issues: the degree to which latent tree learning improves task performance, the degree to which latent tree learning models learn similar grammars across random restarts, the degree to which their grammars match PTB grammar, and the degree to which their grammars appear to follow any recognizable grammatical principles.

We confirm that both types of model succeed at producing usable sentence representations, but find that only the stronger of the two models---that of \citet{choi2017unsupervised}---outperforms either a comparable TreeRNN baseline or a simple LSTM RNN. We find that the grammar of the \citeauthor{choi2017unsupervised} model varies dramatically across random restarts, and tends to agree with PTB grammar with roughly chance accuracy. We do find, though, that the resulting grammar has some regularities, including a preference for shallow trees, a somewhat systematic treatment of negation, and a preference to treat pairs of adjacent words at the edges of a sentence as constituents.

\section{Background}

The work discussed in this paper is closely related to work on \textit{grammar induction}, in which statistical learners attempt to solve the difficult problem of  reconstructing the grammar that generated a corpus of text using only that corpus and, optionally, some heuristics about the nature of the expected grammar. 
Grammar induction in NLP has been widely studied since at least the mid-1990s \citep{chen:1995:ACL,cohen2011unsupervised,hsu2012identifiability}, and builds on a earlier line of more general work in machine learning surveyed in \citet{duda1973pattern} and \citet{fu77syntactic}.
One work in this area, \citet{naseem2011using}, additionally provides some semantic information to the learner, though only as a source of additional guidance, rather than as a primary objective as here. In related work, \citet{gormley-EtAl:2014:P14-1} present a method for jointly training a grammar induction model and a semantic role labeling (SRL) model. They find that the resulting SRL model is more effective than one built on a purely unsupervised grammar induction system, but that using a conventionally trained parser instead yields better SRL performance.

There is also a long history of work on artificial neural network models that build latent hierarchical structures without direct supervision when solving algorithmic problems, including \citet{das1992learning}, \citet{Sun93neural}, and more recently, \citet{joulin2015inferring} and \citet{grefenstette2015learning}.

We are only aware of four prior works on latent tree learning for sentence understanding with neural networks. All four jointly train two model components---a parser based on distributed representations of words and phrases, and a TreeRNN of some kind that uses those parses---but differ in the parsing strategies, TreeRNN parameterizations, and training objective used. 

\begin{figure*}[t]
\centering
\scalebox{0.7}{%
\begin{tikzpicture}
    \def\dx{21pt}
    \def\dy{11pt}
    \def\sy{12*\dy}
    \def\oxb{8*\dx}
    \def\by{0pt}
    \def\ox{0*\oxb}

    \tikzstyle{label}=[text width=35mm,align=center,text height=2mm]    
    \tikzstyle{word}=[text width=35mm,align=center,text height=2mm]    
    \tikzstyle{tracker}=[fill=red!40,text width=15mm,align=center,text height=2mm]
    \tikzstyle{softmax}=[fill=blue!40,text width=15mm,align=center,text height=2mm]
    \tikzstyle{comp}=[fill=green!40,text width=20mm,align=center,text height=2mm]
    \tikzstyle{compoff}=[fill=green!10!black!10,text width=20mm,align=center,text height=2mm]
    \tikzstyle{result}=[line width=1pt,draw=black,text width=15mm,align=center,text height=2mm]    
    \tikzstyle{sbox}=[line width=1pt,draw=black,text width=25mm,align=center,text height=13.3mm]
    \tikzstyle{bbox}=[line width=1pt,draw=black,text width=25mm,align=center,text height=13.3mm]
    \tikzstyle{focus1}=[fill=yellow!40,text width=25mm,align=center,text height=2mm]
    \tikzstyle{focus2}=[fill=yellow!40,text width=25mm,align=center,text height=5.5mm]

    \node[label]  (sl) at (\ox-0.35*\oxb+0*\dx,\by+1*\dy) {buffer};

    \node[focus1] (0bb) at  (\ox+0*\dx,2.5*\dy) {};
    \node[word]  (0b3) at (\ox+0*\dx,\by-0.5*\dy) {};
    \node[word]  (0b2) at (\ox+0*\dx,\by+1.5*\dy) {down};
    \node[word]  (0b1) at (\ox+0*\dx,\by+2.5*\dy) {sat};
    \node[bbox] (0bb) at  (\ox+0*\dx,\by+1.0*\dy) {};
    
    \node[label]  (sl) at (\ox-0.35*\oxb+0*\dx,\sy+0.5*\dy) {stack};
    
    \node[focus2] (0sb) at  (\ox+0*\dx,\sy-0.5*\dy) {};
    \node[word]  (0s1) at (\ox+0*\dx,\sy-1*\dy) {cat};
    \node[word]  (0s2) at (\ox+0*\dx,\sy+0*\dy) {the};
    \node[word]  (0s3) at (\ox+0*\dx,\sy+1*\dy) {};
    \node[sbox] (0sb) at  (\ox+0*\dx,\sy+0.5*\dy) {};
    
    \node[comp] (0c) at  (\ox+0.5*\oxb,\sy-1.5*\dy) {composition};
    
    \node[tracker] (0t) at  (\ox+0*\dx,5*\dy) {tracking};
    \node[softmax] (0sm) at  (\ox+3.25*\dx,6.25*\dy) {transition};
    \node[result] (0so) at  (\ox+3.25*\dx,8.5*\dy) {\textsc{reduce}};
    
    \def\ox{1*\oxb}

    \node[focus1] (1bb) at  (\ox+0*\dx,2.5*\dy) {};
    \node[word]  (1b3) at (\ox+0*\dx,\by-0.5*\dy) {};
    \node[word]  (1b2) at (\ox+0*\dx,\by+1.5*\dy) {down};
    \node[word]  (1b1) at (\ox+0*\dx,\by+2.5*\dy) {sat};
    \node[bbox] (1bb) at  (\ox+0*\dx,\by+1*\dy) {};
    
    \node[focus2] (1sb) at  (\ox+0*\dx,\sy-0.5*\dy) {};
    \node[word]  (1s1) at (\ox+0*\dx,\sy-1*\dy) {the cat};
    \node[word]  (1s2) at (\ox+0*\dx,\sy+0*\dy) {};
    \node[word]  (1s3) at (\ox+0*\dx,\sy+1*\dy) {};
    \node[sbox] (1sb) at  (\ox+0*\dx,\sy+0.5*\dy) {};
    
    \node[compoff] (1c) at  (\ox+0.5*\oxb,\sy-1.5*\dy) {composition};
    
    \node[tracker] (1t) at  (\ox+0*\dx,5*\dy) {tracking};
    \node[softmax] (1sm) at  (\ox+3.25*\dx,6.25*\dy) {transition};
    \node[result] (1so) at  (\ox+3.25*\dx,8.5*\dy) {\textsc{shift}};
     
    \def\ox{2*\oxb}

    \node[focus1] (2bb) at  (\ox+0*\dx,2.5*\dy) {};
    \node[word]  (2b3) at (\ox+0*\dx,\by-0.5*\dy) {};
    \node[word]  (2b2) at (\ox+0*\dx,\by+1.5*\dy) {};
    \node[word]  (2b1) at (\ox+0*\dx,\by+2.5*\dy) {down};
    \node[bbox] (2bb) at  (\ox+0*\dx,\by+1*\dy) {};
    
    \node[focus2] (2sb) at  (\ox+0*\dx,\sy-0.5*\dy) {};
    \node[word]  (2s1) at (\ox+0*\dx,\sy-1*\dy) {sat};
    \node[word]  (2s2) at (\ox+0*\dx,\sy+0*\dy) {the cat};
    \node[word]  (2s3) at (\ox+0*\dx,\sy+1*\dy) {};
    \node[sbox] (2sb) at  (\ox+0*\dx,\sy+0.5*\dy) {};
   
    \node[tracker] (2t) at  (\ox+0*\dx,5*\dy) {tracking};

    \pgfsetarrowsend{latex}
    \tikzstyle{fwd} = [draw=black, line width=1pt]
    \tikzstyle{gated} = [draw=black!50, line width=0.8pt]

    \draw [fwd] (0sb) -- (0t);
    \draw [fwd] (0bb) -- (0t);
    \draw [fwd] (0t) -- (0sm);
    \draw [fwd] (0sm) -- (0so);
    \draw [fwd] (0sb) -- (0c);
    
    \draw [fwd] (0t) -- (1t);
    \draw [fwd] (0t) to[out=50,in=-175] (0c);
    \draw [fwd] (0sb) -- (1sb);
    \draw [fwd] (0bb) -- (1bb);
    \draw [fwd] (0so) to[out=5,in=-140] (1sb);
    \draw [fwd] (0so) to[out=-5,in=170] (1bb);
    \draw [gated] (0bb) to[out=15,in=-125] (1sb);
    \draw [fwd] (0c) -- (1sb);

    \draw [fwd] (1sb) -- (1t);
    \draw [fwd] (1bb) -- (1t);
    \draw [fwd] (1t) -- (1sm);
    \draw [fwd] (1sm) -- (1so);
    \draw [fwd] (1sb) -- (1c);
    
    \draw [fwd] (1t) -- (2t);
    \draw [fwd] (1t) to[out=50,in=-175] (1c);
    \draw [fwd] (1sb) -- (2sb);
    \draw [fwd] (1bb) -- (2bb);
    \draw [fwd] (1so) to[out=5,in=-140] (2sb);
    \draw [fwd] (1so) to[out=-5,in=170] (2bb);
    \draw [fwd] (1bb) to[out=15,in=-125] (2sb);
    \draw [gated] (1c) -- (2sb);

    \draw [fwd] (2sb) -- (2t);
    \draw [fwd] (2bb) -- (2t);

  \end{tikzpicture}}
  
 \caption{\label{fig:spinn} The SPINN model unrolled for two transitions during the processing of the sentence \textit{the cat sat down}, 
reproduced from \protect\citet{bowman2016spinn}. The model parses a sentence by selecting a sequence of \textsc{shift} and \textsc{reduce} transitions using the \textit{transition} classifier (shown in blue) and simultaneously uses the resulting parse to build a vector representation of the sentence by using a TreeLSTM \textit{composition} function (shown in green) during \textsc{reduce} transitions. The \textit{tracking LSTM} (shown in red) maintains a summary of the state of the model which is used in both parsing and composition. The process continues until the buffer is empty and only one representation is left on the stack.}
\end{figure*}
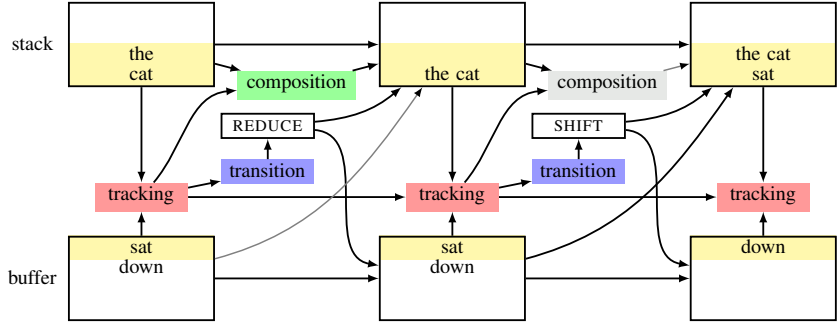

\citet{socher2011semi} present the first neural network model that we are aware of that use the same learned representations to both parse a sentence and---using the resulting parse in a TreeRNN---perform sentence-level classification. They use a plain TreeRNN and a simple parser that scores pairs of adjacent words and phrases and merges the highest-scoring pair. They train their model on a sentiment objective, but rather than training the parsing component on that objective as well, they use a combination of an auxiliary autoencoding objective and a nonparametric scoring function to parse. While this work shows good results on sentiment, it does not feature any evaluation of the induced trees, either through direct analysis nor through comparison with any sentiment baseline that uses trees from a conventionally-trained parser. \citet{bowman2016spinn} introduce an efficient, batchable architecture for doing this---the Shift-reduce Parser-Interpreter Neural Network (SPINN; Figure~\ref{fig:spinn})---which is adapted by \citet{yogatama2016learning} for latent tree learning and used in this work.

The remaining three models all use TreeLSTMs \citep{tai2015improved}, and all train and evaluate both components on a shared semantic objective. All three use the task of recognizing textual entailment on the SNLI corpus \citep{bowman-EtAl:2015:EMNLP} as one such objective. The models differ from one another primarily in the ways in which they use this task objective to train their parsing components, and in the structures of those components.

\citet{yogatama2016learning} present a model (which we call RL-SPINN) that is identical to SPINN at test time, but uses the REINFORCE algorithm \citep{williams1992simple} at training time to compute gradients for the transition classification function, which produces discrete decisions and does not otherwise receive gradients through backpropagation. Surprisingly, and in contrast to \citeauthor{gormley-EtAl:2014:P14-1}, they find that a small 100D instance of this RL-SPINN model performs somewhat better on several text classification tasks than an otherwise-identical model which is explicitly trained to parse.

In unpublished work, \citet{maillard2017jointly} present a model which explicitly computes $O(N^2)$ possible tree nodes for $N$ words, and uses a soft gating strategy to approximately select valid combinations of these nodes that form a tree. This model is trainable entirely using backpropagation, and a 100D instance of the model performs slightly better than RL-SPINN on SNLI.

\citet{choi2017unsupervised} present a model (which we call ST-Gumbel) that uses a similar data structure and gating strategy to \citeauthor{maillard2017jointly}, but which uses the Straight-Through Gumbel-Softmax estimator \citep{jang2016categorical}. This allows them to use a hard categorical gating function, so that their output sentence vector is computed according to a single tree, rather than a gated combination of partial trees as in \citeauthor{maillard2017jointly}. They report substantial gains in both speed and accuracy over \citeauthor{maillard2017jointly} and \citeauthor{yogatama2016learning} on SNLI.
            
Several models \citep{naradowsky:2012:EMNLP,kim2017structured,liu2017learning,munkhdalai-yu:2017:EACLlong2} have also been proposed that can induce latent dependency trees over text using mechanisms like attention, but do not propagate information up the trees as in typical compositional models. Other models like that of \citet{chung2016hierarchical} induce and use latent trees or tree-like structures, but constrain these structures to be of a low fixed depth.       

Other past work has also investigated the degree to which neural network models for non-syntactic tasks \textit{implicitly} learn syntactic information. Recent highlights from this work include \citet{linzen2016assessing} on language modeling and \citet{P17-1080} on translation. Also on translation, \citet{neubig2012} and \citet{denero:2011:EMNLP} present methods that use aligned sentences from bilingual parallel text to learn a binary constituency parser for use in word reordering. 
These two papers do not evaluate these parsers on typical parsing metrics, but find that the parsers support word reorderings that in turn yield improvements in translation quality, suggesting that they do capture some notion of syntactic constituency.

\section{Models and Methods} \label{modelsandmethods}

This paper investigates the behavior of two models: RL-SPINN and ST-Gumbel. Both have been shown to outperform similar models based on supervised parsing, and the two represent substantially different approaches to latent tree learning. 

\paragraph{SPINN Variants} Three of our of our baselines and one of the two latent tree learning models are based on the SPINN architecture of \citet{bowman2016spinn}. Figure \ref{fig:spinn} shows and describes the architecture.

In the base SPINN model, all model components are used, and the transition classifier is trained on binarized Penn Treebank-style parses from the Stanford PCFG Parser \citep{klein2003accurate}, which are included with SNLI and MultiNLI. These binary-branching parse trees are converted to \textsc{shift/reduce} sequences for use in the model through a simple reversible transformation.

RL-SPINN, based on the \textit{unsupervised syntax} model of \citet{yogatama2016learning}, is architecturally equivalent to SPINN, but its transition classifier is optimized for MultiNLI classification accuracy, rather than any parsing-related loss. Because this component produces discrete decisions, the REINFORCE algorithm (with the standard exponential moving average baseline) is used to supply gradients for it. We explored several alternative baseline strategies---including parametric value networks and strategies based on greedy decoding---as well as additional strategies for increasing exploration. We also thoroughly tuned the relevant hyperparameter values for each alternative. In all of these experiments, we found that a standard implementation with the exponential moving average baseline produces accuracy no worse than any readily available alternative.

We also evaluate two other variants of SPINN as baselines. In SPINN-NC (for No Connection from tracking to composition), the connection from the tracking LSTM to the composition function is severed. This weakens the model, but makes it exactly equivalent to a plain TreeLSTM---it will produce the exact same vector that a TreeLSTM with the same composition function would have produced for the tree that the transition classifier implicitly produces. This model serves as a maximally comparable baseline for the ST-Gumbel model, which also performs composition using a standard TreeLSTM in forward-propagation.

SPINN-PI-NT (for Parsed Input, No Tracking) removes the tracking LSTM, as well as the two components that depend on it: the tracking-composition connection and the transition decision function. As such, it cannot produce its own parse trees and must rely on trees from the input data. We include this in our comparison to understand the degree to which training a parser, rather than using a higher-quality off-the-shelf parser, impacts performance on our semantic task. 

\begin{figure}[t]
\centering\scalebox{0.7}{%
    \begin{tikzpicture}[
        node distance=7em,
        every node/.style = {shape=rectangle, draw,
                             minimum height=1.5em, line width=1pt,
                             align=center, text height=2mm},
        every edge/.append style = {line width=1pt}]
    \node[text width=15mm,              fill=yellow!40] (f1) {the cat};
    \node[right of=f1, text width=15mm, fill=yellow!40] (f2) {sat};
    \node[right of=f2, text width=15mm, fill=yellow!40] (f3) {down};
    \node[left of=f1, node distance=7em, draw=none] (flabel) {\scalebox{.8}{layer 1}};
    
    \node[below=2em of f1, text width=15mm, fill=green!40] (c1) {the cat};
    \node[below=2em of f2, text width=15mm, dashed, fill=green!10!black!10] (c2) {cat sat};
    \node[below=2em of f3, text width=15mm, dashed, fill=green!10!black!10] (c3) {sat down};
    \node[right of=c3, node distance=7em, draw=none, align=left] (clabel) {\scalebox{.6}{}};
    
    \node[below=1.5em of c1, text width=15mm, xshift=-3em, fill=yellow!40] (l1) {the};
    \node[below=1.5em of c1, text width=15mm, xshift=3em, fill=yellow!40] (l2) {cat};
    \node[below=1.5em of c2, text width=15mm, xshift=3em, fill=yellow!40] (l3) {sat};
    \node[below=1.5em of c3, text width=15mm, xshift=3em, fill=yellow!40] (l4) {down};
    \node[left of=l1, node distance=4em, draw=none] (llabel) {\scalebox{.8}{layer 0}};
    
    \node[left of=c1, shape=circle, fill=blue!40] (q) {\small $\mathbf{q}$};
    \draw [->, line width=1pt, color=gray, dashed]
        (q.north)
        -- ([yshift=1em]q.north)
        -- ([xshift=6em, yshift=1em]q.north)
        -- ([xshift=-0.3em]c1.north);
    \draw [<-, line width=1pt, color=gray, dashed]
        ([xshift=0.2em]q.north)    
        -- ([xshift=0.2em,yshift=0.8em]q.north)
        -- ([xshift=5.8em, yshift=0.8em]q.north)
        -- ([xshift=-0.1em]c1);
    \node[left of=c1, node distance=2.6em, yshift=1.3em, draw=none] {\scalebox{.7}{$v_1=0.5$}};
    \draw [->, line width=1pt, color=gray, dashed]
        (q.north)
        -- ([yshift=1em]q.north)
        -- ([xshift=13em, yshift=1em]q.north)
        -- ([xshift=-0.3em]c2.north);
    \draw [<-, line width=1pt, color=gray, dashed]
        ([xshift=0.2em]q.north)    
        -- ([xshift=0.2em,yshift=0.8em]q.north)
        -- ([xshift=12.8em, yshift=0.8em]q.north)
        -- ([xshift=-0.1em]c2);
    \node[left of=c2, node distance=2.6em, yshift=1.3em, draw=none] {\scalebox{.7}{$v_2=0.1$}};
    \draw [->, line width=1pt, color=gray, dashed]
        (q.north)
        -- ([yshift=1em]q.north)
        -- ([xshift=20em, yshift=1em]q.north)
        -- ([xshift=-0.3em]c3.north);
    \draw [<-, line width=1pt, color=gray, dashed]
        ([xshift=0.2em]q.north)    
        -- ([xshift=0.2em,yshift=0.8em]q.north)
        -- ([xshift=19.8em, yshift=0.8em]q.north)
        -- ([xshift=-0.1em]c3);
    \node[left of=c3, node distance=2.6em, yshift=1.3em, draw=none] {\scalebox{.7}{$v_3=0.4$}};

    \draw [->] (c1) edge (f1);
    \draw [->] (l1) edge (c1);
    \draw [->] (l2) edge (c1);
    \draw [dashed,->] (l2) edge (c2);
    \draw [dashed,->] (l3) edge (c2);
    \draw [dashed,->] (l3) edge (c3);
    \draw [dashed,->] (l4) edge (c3);
    
    \draw [->, bend right=30] (l3) edge (f2);
    \draw [->, bend right=30] (l4) edge (f3);
    
    \end{tikzpicture}}
 \caption{\label{fig:gumbel} The ST-Gumbel model in its first step of processing the sentence \textit{the cat sat down}, based on a figure by \citet{choi2017unsupervised}, used with permission. The model first computes a composed representation for every pair of adjacent words or phrases (shown with dotted lines), assigns each of these a score $v_i$, and then uses these to parameterize a distribution from which the discrete random variable $q$ is sampled. The sampled value then determines which representation is preserved for the next layer.}
\end{figure}
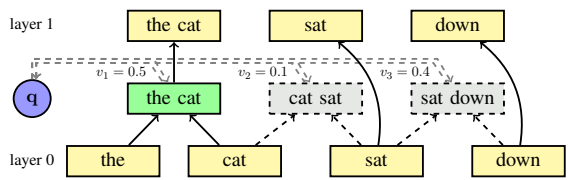

\paragraph{ST-Gumbel}
The ST-Gumbel model was developed by \citet{choi2017unsupervised} and is shown in Figure \ref{fig:gumbel}. The model takes a sequence of $N
- 1$ steps to build a tree over $N$ words. At every step, every possible pair of adjacent words or phrase vectors in the partial tree is given to a TreeLSTM composition function to produce a new candidate phrase vector. A simple learned scoring function then selects the best of these candidates, which forms a constituent node in the tree and replaces its two children in the list of nodes that are available to compose. This repeats until only two nodes remain, at which point they are composed and the tree is complete. This exhaustive search increases the computational complexity of the model over (RL-)SPINN, but also allows the model to perform a form of easy-first parsing, making it easier for the model to explore the space of possible parsing strategies.

Though the scoring function yields discrete decisions, the \citeauthor{jang2016categorical} Straight-Through Gumbel-Softmax estimator makes it possible to nonetheless efficiently compute an approximate gradient for the full model without the need for relatively brittle policy gradient  techniques like REINFORCE.

\paragraph{Other Baselines}
We also train three baselines that do not depend on a parser. The first is a unidirectional LSTM RNN over the embedded tokens. The second is a version of SPINN-PI-NT that is supplied sequences of randomly sampled legal transitions (corresponding to random parse trees), rather than the output of a parser. The third is also a version of SPINN-PI-NT, and receives transition sequences corresponding to maximally-shallow, approximately-balanced parse trees based on the ``full binary'' trees used in \citet{munkhdalai-yu:2017:EACLlong1}. 

\paragraph{Data} To ensure that we are able to roughly reproduce the results reported by \citeauthor{yogatama2016learning} and \citeauthor{choi2017unsupervised}, we conduct an initial set of experiments on the Stanford NLI corpus of \citet{bowman-EtAl:2015:EMNLP}. Our primary experiments use the newer Multi-Genre Natural Language Inference Corpus \citep[MultiNLI;][]{williams2017broad}. MultiNLI is a 433k-example textual entailment dataset created in the style of SNLI, but with a more diverse range of source texts and longer, more complex sentences, which we expect will encourage the models to produce more consistent and interpretable trees than they otherwise might. Following \citeauthor{williams2017broad}, we train on the combination of MultiNLI and SNLI in these experiments (yielding just under 1M training examples) and evaluate on MultiNLI (using the \textit{matched}  development and test sets). 

We also evaluate trained models on the full Wall Street Journal section of the Penn Treebank, a seminal corpus of manually-constructed constituency parses, which introduced the parsing standard used in this work. Because the models under study produce and consume binary-branching constituency trees without labels (and because such trees are already included with SNLI and MultiNLI), we use the Stanford Parser's \texttt{CollapseUnaryTransformer} and \texttt{TreeBinarizer} tools to convert these Penn Treebank Trees to this form.
        
\paragraph{Sentence Pair Classification}
Because our textual entailment task requires a model to classify \textit{pairs} of sentences, but the models under study produce vectors for single sentences, we concatenate the two sentence vectors, their difference, and their elementwise product \citep[following][]{P16-2022}, and feed the result into a 1024D ReLU layer to produce a representation for the sentence pair. This representation is fed into a three-way softmax classifier that selects one of the labels \textit{entailment}, \textit{neutral}, and \textit{contradiction} for the pair.

\paragraph{Additional Details} We implement all models in PyTorch 0.2. We closely follow the original Theano code for SPINN in our implementation, and we incorporate source code provided by \citeauthor{choi2017unsupervised}  for the core parsing data structures and sampling mechanism of the ST-Gumbel model. Our code, saved models, and model output are available on GitHub.\footnote{\protect\url{https://github.com/nyu-mll/spinn/tree/is-it-syntax-release}}

We use GloVe vectors to represent words \citep[standard 300D, 840B word package, without fine tuning;][]{pennington2014glove}, and feed them into a single-layer $2 \times 300$D bi-directional GRU RNN (based on the \textit{leaf LSTM} of \citeauthor{choi2017unsupervised}) to give the models access to local context information when making parsing decisions. To understand the impact of this component, we follow \citeauthor{choi2017unsupervised} in also training each model with the leaf GRU replaced with a simpler context-insensitive input encoder that simply multiplies each GloVe vector by a matrix. We find that these models perform best when the temperature of the ST-Gumbel distribution is a trained parameter, rather than fixed at 1.0 as in \citeauthor{choi2017unsupervised}.

We use L2 regularization and apply dropout \citep{srivastava2014dropout} to the input of the 1024D sentence pair combination layer. We train all models using the Adam optimizer \citep{kingma2014adam}. For hyperparameters for which no obvious default value exists---the L2 and dropout parameters, the relative weighting of the gradients from REINFORCE in RL-SPINN, the starting learning rate, and the size of the tracking LSTM state in SPINN---we heuristically select ranges in which usable values can be found (focusing on MultiNLI development set performance), and then randomly sample values from those ranges. We train each model five times using different samples from those ranges and different random initializations for model parameters. We use early stopping based on development set performance with all models.

\begin{table}[t!]
\small \centering
\begin{tabular}{lcc}
\toprule
\bf Model & \bf SNLI & \bf MNLI\\
\midrule
\multicolumn{3}{c}{Prior Work: Baselines} \\
\midrule
100D LSTM (Yogatama) & 80.2 & -- \\
100D TreeLSTM (Yogatama) & 78.5 & -- \\
300D SPINN (Bowman) & 83.2 & -- \\
300D SPINN-PI-NT (Bowman) & 80.9 & -- \\
300D BiLSTM (Williams) & 81.5 & 67.5 \\
\midrule
\multicolumn{3}{c}{Prior Work: Latent Tree Learning} \\
\midrule
100D RL-SPINN (Yogatama) & 80.5 & -- \\
100D Soft Gating (Maillard) & 81.6 & -- \\
100D ST-Gumbel (Choi) & 81.9 & -- \\
\hspace{1em} w/o Leaf LSTM & 80.2 & -- \\
300D ST-Gumbel (Choi) & 84.6 & -- \\
\hspace{1em} w/o Leaf LSTM & 82.2 & -- \\
600D ST-Gumbel (Choi) & \textbf{85.4} & -- \\
\midrule
\multicolumn{3}{c}{This Work: Baselines} \\
\midrule
300D LSTM & 82.6 & 69.1  \\
300D SPINN & 81.9 & 66.9  \\
\hspace{1em} w/o Leaf GRU & 82.2 & 67.5 \\ 
300D SPINN-NC & 81.6 & 68.1 \\
\hspace{1em} w/o Leaf GRU & 82.4 & 67.8 \\ 
300D SPINN-PI-NT & 81.9 & 68.2  \\
\hspace{1em} w/o Leaf GRU & 81.7 & 67.6 \\ 
300D SPINN-PI-NT (Rand.) & 81.6 & 68.0  \\
\hspace{1em} w/o Leaf GRU & 80.4 & 66.2 \\ 
300D SPINN-PI-NT (Bal.) & 81.9 & 68.2 \\
\hspace{1em} w/o Leaf GRU & 81.3 & 66.5 \\
\midrule
\multicolumn{3}{c}{This Work: Latent Tree Learning} \\
\midrule
300D ST-Gumbel & 83.3 & \textbf{69.5}  \\
\hspace{1em} w/o Leaf GRU & 83.7 & 67.5 \\ 
300D RL-SPINN & 81.7 & 67.3  \\
\hspace{1em} w/o Leaf GRU & 82.3 & 67.4 \\ 
\bottomrule
\end{tabular}  
\caption{\label{tab:acc}Test set results. Our implementations of SPINN and RL-SPINN differ only in how they are trained to parse, and our implementations of SPINN-NC and ST-Gumbel also differ only in how they are trained to parse. SPINN-PI-NT includes no tracking or parsing component and instead uses externally provided Stanford Parser trees, random trees (Rand.) or balanced trees (Bal.), as described in Section~\ref{modelsandmethods}.}
\end{table}

\begin{table}[t!]
\small
\centering
\begin{tabular}{lccc}
\toprule
 & \multicolumn{2}{c}{\bf Dev. Acc. (\%)} & \bf Self \\
\bf Model & $\boldsymbol{\mu~(\sigma)}$ & \bf max & \bf F1 \\
\midrule
300D SPINN & 67.1 (1.0) & 68.3 & 71.5  \\
\hspace{1em} w/o Leaf GRU & 68.3 (0.4) & 68.9 & 67.7 \\ 
300D SPINN-NC & 67.7 (0.7) & 68.5 & 73.2  \\
\hspace{1em} w/o Leaf GRU & 68.7 (0.4) & 69.1 & 67.6 \\ 
\midrule
300D ST-Gumbel & 69.1 (0.7) & \textbf{70.0} & 49.9  \\
\hspace{1em} w/o Leaf GRU & 63.5 (6.5) & 68.1 & 41.2 \\ 
300D RL-SPINN & 67.4 (1.6) & 68.6 & 92.7  \\
\hspace{1em} w/o Leaf GRU & 68.8 (0.4) & 69.3 & \textbf{98.5} \\ 
\midrule
Random Trees & -- & -- & 32.6 \\
Balanced Trees & -- & -- & 100.0 \\
\bottomrule
\end{tabular}  
\caption{\label{tab:selff1}\textit{Dev.~Acc.}~shows MultiNLI development set accuracy across the five runs (expressed as mean, standard deviation, and maximum). \textit{Self F1} shows how well each of the five models agrees in its parsing decisions with the other four.
}
\end{table}

\begin{table*}[t!]
\small
\centering
\begin{tabular}{lccccccc}
\toprule
 & \multicolumn{2}{c}{\bf F1} & \multicolumn{4}{c}{\bf Accuracy} \\
\bf Model & $\boldsymbol{\mu~(\sigma)}$ & \bf max & \bf ADJP & \bf NP & \bf PP & \bf INTJ  \\
\midrule
300D SPINN &    53.4 (5.6) & \bf 59.6  & 29.3 & 48.3 & 39.0  & 57.1 \\
\hspace{1em} w/o Leaf GRU & 44.4 (2.7) & 47.7  & 25.2 & 39.2 & 30.8  & 60.0  \\
300D SPINN-NC & 54.3 (3.0) & 58.7  & \bf 31.8 & \bf 49.3 & \bf 39.3 &  58.2  \\
\hspace{1em} w/o Leaf GRU & 45.3 (3.0) & 49.2  & 27.1 & 39.9 & 31.4  & \bf 60.6  \\
\midrule 
300D ST-Gumbel &  \it 19.0 (1.0) & \it 20.1  & \it 15.6 & \it 18.8 & \it \it 9.9 &  59.4  \\
\hspace{1em} w/o Leaf GRU & 22.8 (1.6) & 25.0  & 18.9 & 24.1 & \it 14.2 & 51.8  \\
300D RL-SPINN & \it 13.2 (0.0) & \it 13.2  & \it 1.7 & \it 10.8 & \it 4.6 & 50.6  \\
\hspace{1em} w/o Leaf GRU & \it 13.1 (0.1) & \it 13.2  & \it 1.6 & \it 10.9 & \it 4.6 & 50.0 \\
\midrule 
Random Trees &  21.3 (0.0) & 21.4  & 17.4 & 22.3 & 16.0 & 40.4 \\ 
Balanced Trees &  21.3 (0.0) & 21.3  & 22.1 & \textit{20.2} & \textit{9.3} & 55.9 \\
\bottomrule
\end{tabular} 
\caption{\label{tab:ptb}Results on the full Wall Street Journal section of the Penn Treebank. The \textit{F1} columns represent overall unlabeled binary F1. The \textit{Accuracy} columns represent the fraction of ground truth constituents of a given type that correspond to constituents in the model parses. Italics mark results that are worse than the random baseline.}
\end{table*}

\section{Does latent tree learning help sentence understanding?}

Table \ref{tab:acc} shows the accuracy of all models on two test sets: SNLI (training on SNLI only, for comparison with prior work), and MultiNLI (training on both datasets). Each figure represents the accuracy of the best run, selected using the development set, of five runs with different random initializations and hyperparameter values.

Our LSTM baseline is strikingly effective, and matches or exceeds the performance of all of our PTB grammar-based tree-structured models on both SNLI and MultiNLI. This contradicts the primary result of \citet{bowman2016spinn}, and suggests that there is little value in using the correct \textit{syntactic} structure for a sentence to guide neural network composition, at least in the context of the TreeLSTM composition function and the NLI task.

We do, however, reproduce the key result of \citeauthor{choi2017unsupervised} on both datasets. Their ST-Gumbel model, which receives no syntactic information at training time, outperforms SPINN-NC, which performs composition in an identical manner but is explicitly trained to parse, and also outperforms the LSTM baseline. This suggests that the learned latent trees are helpful in the construction of semantic representations for sentences, whether or not they resemble conventional parse trees.

Our results with RL-SPINN are more equivocal. That model matches, but does not beat, the performance of the full SPINN model, which is equivalent except that it is trained to parse. However, our implementation of RL-SPINN outperforms \citeauthor{yogatama2016learning}'s (lower-dimensional) implementation by a substantial margin. The impact of the leaf GRU is sometimes substantial, but the direction of its effect is not consistent. 

Our results with SPINN-PI-NT are not substantially better than those with any other model, suggesting the relatively simple greedy parsing strategies used by the other models are not a major limiting factor in their performance. Balanced trees consistently outperform randomly sampled transitions (albeit by a small margin), yet perform worse than ST-Gumbel even though ST-Gumbel uses very shallow trees as well. Similarly, RL-SPINN depends on mostly left-branching binary parse trees, but is outperformed by a forward LSTM. Structure is important, but there are differences between the architectures of compositional models worth investigating in future work.

None of our latent tree learning models reach the state of the art on either task, but all are comparable in both absolute and relative performance to other published results, suggesting that we have trained reasonable examples of latent tree learning models and can draw informative conclusions by studying the behaviors of these models. 

\section{Are these models consistent?}

If it were the case that a latent tree learning model outperforms its baselines by identifying some specific grammar for English that is better than the one used in PTB and the Stanford Parser, then we would expect these models to identify roughly the same grammar across random restarts and minor configuration changes, and to use that grammar to produce consistent task performance.
Table \ref{tab:selff1} shows two measures of consistency for the four models that produce parses, a simple random baseline that produces parses by randomly merging pairs of adjacent words and phrases, and (trivially) the deterministic strategy used in the Balanced Trees runs. 

We first show the variation in \textit{accuracy} on the MultiNLI development set across runs. While one outlier distorts these numbers for ST-Gumbel without the leaf GRU, these figures are roughly equivalent between the latent tree learning models and the baselines, suggesting that these models are not substantially more brittle or more hyperparameter sensitive in their task performance. The second metric shows the \textit{self F1} for each model: the unlabeled binary F1 between the parses produced by two runs of the same model for the MultiNLI development set, averaged out over all possible pairings of different runs. This measures the degree to which the models reliably converge on the same parses, and sheds some light on the behavior of the models. The baseline models show relatively high consistency, with self F1 above 65\%. ST-Gumbel is substantially less consistent, with scores below 50\% but above the 32.6\% random baseline. RL-SPINN appears to be highly consistent, with the runs without the leaf GRU reaching 98.5\% self F1, suggesting that it reliably converges to a specific grammar. However, as we will discuss in later sections, this grammar appears to be trivial. 

\begin{table*}
\small
\centering
\begin{tabular}{lcccccccc}
\toprule
 & \multicolumn{6}{c}{\bf F1 wrt.} & \multicolumn{2}{c}{\bf Macroavg.}\\
 & \multicolumn{2}{c}{\bf Left Branching} & \multicolumn{2}{c}{\bf Right Branching} & \multicolumn{2}{c}{\bf Stanford Parser}  & \multicolumn{2}{c}{\bf Depth}\\
\bf Model & $\boldsymbol{\mu~(\sigma)}$ & \bf max & $\boldsymbol{\mu~(\sigma)}$ & \bf max  & $\boldsymbol{\mu~(\sigma)}$ & \bf max  & $\boldsymbol{\mu~(\sigma)}$ & \bf max \\
\midrule
300D SPINN & 19.3 (0.4) & 19.8 & 36.9 (3.4) & \bf 42.6 & 70.2 (3.6) & \bf 74.5 & 6.2 (0.2) & 6.5 \\
\hspace{1em} w/o Leaf GRU & 21.2 (1.4) & 22.9 & 39.0 (2.6) & 41.4 & 63.5 (1.7) & 65.7 & 6.4 (0.1) & 6.5 \\
300D SPINN-NC & 19.2 (0.4) & 19.7 & 36.2 (1.4) & 38.2 & 70.5 (2.0) & 73.1 & 6.1 (0.0) & 6.1 \\
\hspace{1em} w/o Leaf GRU & 20.6 (0.5) & 21.3 & 38.9 (2.5) & 41.9 & 64.1 (2.7) & 67.1 & 6.3 (0.0) & 6.3 \\
\midrule
300D ST-Gumbel & 32.6 (2.0) & 35.6 & 37.5 (2.4) & 40.3 & 23.7 (0.9) & 25.2 & 4.1 (0.1) & 4.2 \\
\hspace{1em} w/o Leaf GRU & 30.8 (1.2) & 32.3 & 35.6 (3.3) & 39.9 & 27.5 (1.0) & 29.0 & 4.6 (0.1) & 4.7 \\
300D RL-SPINN & 95.0 (1.4) & 96.6 & 13.5 (1.8) & 15.8 & 18.8 (0.2) & 19.0 & 8.6 (0.0) & \bf 8.6 \\
\hspace{1em} w/o Leaf GRU & 99.1 (0.6) & \bf 99.8 & 10.7 (0.2) & 11.1 & 18.1 (0.1) & 18.2 & 8.6 (0.0) & \bf 8.6 \\
\midrule
\citeauthor{yogatama2016learning} (SNLI) & -- --  & 41.4  &   -- -- & 19.9  & -- --  & 41.7  &  -- -- & -- \\
\midrule
Random Trees & 27.9 (0.1) & 27.9 &  28.0 (0.1) & 28.1 & 27.0 (0.1) & 27.1 & 4.4 (0.0) & 4.4 \\
Balanced Trees & 21.7 (0.0) & 21.7 &  36.8 (0.0) & 36.8 & 21.3 (0.0) & 21.3 & 3.9 (0.0) & 3.9 \\
\bottomrule
\end{tabular} 
\caption{\label{tab:main}\textit{F1 wrt.}~shows F1 scores on the MultiNLI development set with respect to strictly right- and left-branching trees and with respect to the Stanford Parser trees supplied with the corpus; The F1 results from \citeauthor{yogatama2016learning} are reported for SNLI. \textit{Macroavg.~Depth} shows the average across sentences of the average depth of each word in its tree. Each is shown with the mean, standard deviation, and maximum of the metric across the five runs of each model.}
\end{table*}

\section{Do these models learn PTB grammar?}\label{sec:learnptb}

Given that our latent tree learning models are at least somewhat consistent in what they learn, it is reasonable to ask what it is that they learn. We investigate this first quantitatively, then, in the next section, more qualitatively.

Table \ref{tab:ptb} shows parsing performance on the Wall Street Journal sections of PTB for models trained on SNLI and MultiNLI. The baseline models perform fairly poorly in absolute terms, as they are neither well tuned for parse quality nor trained on news text, but the latent tree learning models perform dramatically worse. The ST-Gumbel models perform at or slightly above chance (represented by the random trees results), while the RL-SPINN models perform consistently below chance. These results suggest that these models do not learn grammars that in any way resemble PTB grammar. 

To confirm this, we also show results for individual Penn Treebank nonterminal node types. On common intermediate node types such as ADJP, NP, and PP, the latent tree learning models do not perform substantially better than chance. It is only on a two rare types that any latent tree learning model, or the balanced tree baseline, outperforms random trees by a significant margin: INTJ (interjection, as in \textit{\textbf{Oh no}, he 's...}) and the even rarer LST (list marker, as in \textit{\textbf{1 .} Determine if...}), both of which are generally short and sentence-initial (discussed in more detail in Section~\ref{discussion}).

Next, we turn to the MultiNLI development set for further investigation.
Table \ref{tab:main} shows results on MultiNLI for a wider range of measures. The table shows F1 measured with respect to three different references: automatically generated trivial trees for the corpus that are either strictly left-branching or strictly right-branching, and the PTB-style trees produced by the Stanford Parser for the corpus. We see that the baseline models perform about as well on MultiNLI as on PTB, with scores above 65\%, and that these models produce trees that tend toward right branching rather than left branching. 

The ST-Gumbel models perform only at or slightly above chance on the parsed sentences, and show a similar use of both right- and left-branching structures, with only a slight preference for the more linguistically common right-branching structures. This suggests that they learn grammars that differ quite substantially from PTB grammars, but may share some minor properties. Our implementation of an ST-Gumbel model has F1 scores with respect to left-branching and Stanford Parser trees that are much closer to the ones \citeauthor{yogatama2016learning} report than to the ones we find for our RL-SPINN implementation.

Our RL-SPINN results are unequivocally negative. We find that our models could be tuned to produce trees that are qualitatively similar to those in \citeauthor{yogatama2016learning}. However, in our primary experiments, we tune model hyperparameters with the sole criterion of downstream task performance, and find that the trees from these experiments yield relactively trivial trees, with F1 scores that are much lower than theirs with respect to the Stanford Parser, and much higher with respect to left branching trees (see Table \ref{tab:main}). All runs perform below chance on the parsed sentences, and all have F1 scores over 92\% with respect to the left-branching structures, suggesting that they primarily learn to produce strictly left-branching trees. This trivial strategy, which makes the model roughly equivalent to a sequential RNN, is very easy to learn. In a shift--reduce model like SPINN, the model can simply learn to perform the \textsc{reduce} operation whenever it is possible to do so, regardless of the specific words and phrases being parsed. This can be done by setting a high bias value for this choice in the transition classifier. 

The rightmost column shows another measure of what is learned: the average depth---the length of the path from the root to any given word---of the induced trees. For the baseline models, this value is slightly above the 5.7 value for the Stanford Parser trees. For the RL-SPINN models, this number is predictably much higher, reflecting the very deep and narrow left-branching trees that those models tend to produce. For the ST-Gumbel model, though, this metric is informative: the models consistently produce shallow trees with depth under 5---closer to the balanced trees baseline than to SPINN. This hints at a possible interpretation: While shallower trees may be less informative about the structure of the sentence than real PTB trees, they reduce the number of layers that a word needs to pass through to reach the final classifier, potentially making it easier to learn an effective composition function that faithfully encodes the contents of a sentence. This interpretation is supported by the results of \citet{munkhdalai-yu:2017:EACLlong1}, who show that it is possible to do well on SNLI using a TreeLSTM (with a leaf LSTM) over arbitrarily chosen balanced trees with low depths, and our balanced trees baseline, which approximates this result.

The ST-Gumbel models tend to implement their shallow parsing strategy with a good deal of randomness. They tend to assign near-zero probability ($<10^{-10}$) to many possible compositions, generally those that would result in unnecessarily deep trees, and relatively smooth probabilities (generally $>0.01$) to the remaining options. The trainable temperature parameter for these models generally converged slowly to a value between 1 and 20, and did not fluctuate substantially during training.

\section{Analyzing the Learned Trees}\label{discussion}

In the previous three sections, we have shown that latent tree learning models are able to perform as well or better than models that have access to linguistically principled parse trees at training or test time, but that the grammars that they learn are neither consistent across runs, nor meaningfully similar to PTB grammar. In this section, we investigate the trees  produced by these learned grammars directly to identify whether they capture any recognizable syntactic or semantic phenomena.

The RL-SPINN models create overwhelmingly left-branching trees. We observe few deviations from this pattern, which occur almost exclusively on sentences with fewer than seven words. Given that the self-F1 scores for these models (92.7 and 98.5, Table~\ref{tab:selff1}) are similar to their F1 scores with respect to strictly left-branching trees (95.0 and 99.1, Table~\ref{tab:main}), there is little room for these models to have learned any consistent behavior beside left branching.
 
In some preliminary tuning runs not shown above, we saw models that deviated from this pattern more often, and one that fixated on \textit{right}-branching structures instead, but we find no grammatically interesting patterns in any of these deviant structures. 

The ST-Gumbel models learned substantially more complex grammars, and we focus on these for the remainder of the section. We discuss three model behaviors which yield linguistically implausible constituents. The first two highlight settings where the ST-Gumbel model is consistent where it shouldn't be, and the third highlights a setting in which it is worryingly inconsistent. The models' treatment of these three phenomena and our observation of these models' behavior more broadly suggest that the models do not produce trees that follow any recognizable semantic or syntactic principles. 

\paragraph{Initial and Final Two-Word Constituents}

The shallow trees produced by the ST-Gumbel models generally contain more constituents comprised of two words (rather than a single word combined with a phrase) than appear in the reference parses. This behavior is especially pronounced at the edges of sentences, where the models frequently treat the first two words and the last two words as constituents. Since this behavior does not correspond to any grammatical phenomenon known to these authors, it likely stems from some unknown bias within the model design.
 
\begin{figure}[t]\centering
	\scalebox{0.8}{
		\begin{forest}
			shape=coordinate,
			where n children=0{
				tier=word
			}{},
			nice empty nodes
			[ [ [\textbf{The}] [\textbf{grammar}] ] [ [was] [ [great] [$.$] ] ] ]
	\end{forest}}~~
	\scalebox{0.8}{
		\begin{forest}
			shape=coordinate,
			where n children=0{
				tier=word
			}{},
			nice empty nodes
			[ [ [ [\textbf{He}] [\textbf{shot}] ] [ [his] [gun] ] ] [ [ [at] [the] ] [ [man] [$.$] ] ] ]
	\end{forest}}~~
\\ \vspace{0.5em}
	\scalebox{0.8}{
	\begin{forest}
	 shape=coordinate,
	 where n children=0{
	  tier=word 
	  			}{},
	 nice empty nodes
	[ [ [ [\textbf{Kings}] [\textbf{frequently}] ] [ [founded] [orders] ] ] [ [ [that] [can] ] [ [ [ [still] [be] ] [found] ] [ [today] [$.$] ] ] ] ] 
	\end{forest}} 
	\caption{The ST-Gumbel models often form constituents from the first two words of sentences.}\label{fig:firsttwo} 
\end{figure}
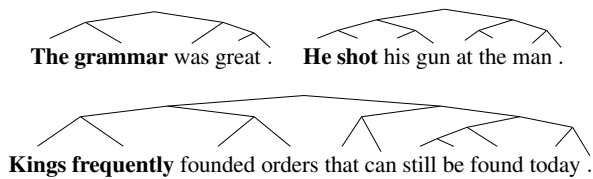


These models parse the first two words of a sentence into a constituent at rates well above both the 50\% rate seen in random and balanced parses\footnote{The balanced parses are right-aligned, following Munkhdalai and Yu; they parse the first two words as a constituent in about 50\% of cases, but the final two words in all cases.} and the 27.7\% rate seen with the SPINN models. This strategy appeares in 77.4\% of the model's parses with the leaf GRU and 64.1\% without. While it was consistently discovered across all of our runs of ST-Gumbel models with the leaf GRU, it was discovered less frequently across restarts for runs without, which do not have direct access to linear position information. We observe that the models combine the final two words in each sentence at similar rates. 

While merging the final two words of a sentence nearly always results in a meaningless constituent containing a period or punctuation mark, merging the first two words can produce reasonable parses. This strategy is reasonable, for example, in sentences that begin with a determiner and a noun (Figure \ref{fig:firsttwo}, top left). However, combining the first two words in sentences that start with adverbials, proper names, bare plurals, or noun phrases with multiple modifiers will generally result in meaningless constituents like \textit{Kings frequently} (Figure \ref{fig:firsttwo}, bottom). 

Combining the first two words of a sentence also often results in more subtly unorthodox trees---like the one in the top right of Figure \ref{fig:firsttwo}---that combine a verb with its subject rather than its object. This contrasts with some mainstream syntactic theories (\citealt{adger,sportiche2013}), which generally take the object and the verb of a sentence to form a constituent for three reasons: Taking the top right sentence in Figure~\ref{fig:firsttwo} as an example, (i) we can replace it with a new constituent of the same type without changing the surrounding sentence structure, as in \textit{he \textbf{did so}}, (ii) it can stand alone as an answer to a question like \textit{what did he do?}, and (iii) it can be omitted in otherwise-repetitive sentences like \textit{He shot his gun, but Bill didn't \underline{~~~}}. 

\paragraph{Negation}

The ST-Gumbel models also tend to learn a systematic and superficially reasonable strategy for negation: they pair any negation word (e.g., \textit{not, n't, no, none}) with the word that immediately follows it. Random parses only form these constituents in 34\% of the sentences, and balanced parses only do so in 50\%, but the ST-Gumbel models with the leaf GRU do so about 67\% of the time and consistently across runs, while those without the leaf GRU do so less consistently, but over 90\% of the time in some runs.

\begin{figure}
	\centering
	\scalebox{0.8}{
		\begin{forest}
			shape=coordinate,
			where n children=0{
				tier=word
			}{},
			nice empty nodes
			[ [It] [ ['s] [ [ [\textbf{not}] [\textbf{predictable}] ] [$.$] ] ] ]
	\end{forest} }
	\scalebox{0.8}{
	\begin{forest}
		shape=coordinate,
		where n children=0{
			tier=word
		}{},
		nice empty nodes
		[ [It] [ [ [ [\textbf{'s}] [\textbf{not}] ] [predictable] ] [$.$] ] ]
\end{forest}}\\\vspace{0.5em}
	\scalebox{0.8}{
	\begin{forest}
		shape=coordinate,
		where n children=0{
			tier=word
		}{},
		nice empty nodes
		[ [He] [ [is] [ [ [ [ [\textbf{not}] [\textbf{at}] ] [all] ] [sure] ] [$.$] ] ] ]
\end{forest}}~~
	\scalebox{0.8}{
	\begin{forest}
		shape=coordinate,
		where n children=0{
			tier=word
		}{},
		nice empty nodes
		[ [He] [ [ [ [\textbf{is}] [\textbf{not}] ] [ [ [at] [all] ] [sure] ] ] [$.$] ] ]
\end{forest}}

	\caption{Left: ST-Gumbel models reliably attach negation words to the words directly following them. Right: Stanford Parser trees for comparison. 
	}\label{fig:negation}
\end{figure}
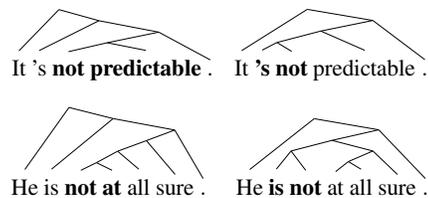

This strategy is effective when the negation word is meant to modify a single other word to its right, as in Figure \ref{fig:negation}, top left sub-figure, but this is frequently not the case. In Figure \ref{fig:negation}, bottom left, although the model creates the potentially reasonable constituent, \textit{not at all}, it also combines \textit{not} with the preposition \textit{at} to form a constituent with no clear interpretation (or, at best, an incredibly bizarre one). Further, combining \textit{not} with \textit{at} goes contra the syntactic observation that prepositional phrases can generally move along with the following noun phrases as a constituent (as in semantically comparably sentences like, \textit{He is not sure \textbf{at all}.}).  

\paragraph{Function Words and Modifiers}

\begin{figure}
	\centering
	\scalebox{0.67}{
		\begin{forest}
			shape=coordinate,
			where n children=0{
				tier=word
			}{},
			nice empty nodes 
			[ [ [ [We] [briefly] ] [ [discussed] [\textbf{the}] ] ] [ [\textbf{Nazi}] [\textbf{angle}] ] ]
	\end{forest}}
	\scalebox{0.67}{
		\begin{forest}
			shape=coordinate,
			where n children=0{
				tier=word
			}{},
			nice empty nodes
			[ [We] [ [briefly] [ [discussed] [ [\textbf{the}] [ [\textbf{Nazi}] [\textbf{angle}] ] ] ] ] ]
	\end{forest}}\\\vspace{0.5em}
	\scalebox{0.67}{
		\begin{forest}
			shape=coordinate,
			where n children=0{
				tier=word
			}{},
			nice empty nodes
			[ [ [ [ [The] [students] ] [reacted] ] [\textbf{with}] ] [ [\textbf{horror}] [$.$] ] ] 
	\end{forest}}
	\scalebox{0.67}{
		\begin{forest}
			shape=coordinate,
			where n children=0{
				tier=word
			}{},
			nice empty nodes
			[ [ [The] [students] ] [ [ [reacted] [ [\textbf{with}] [\textbf{horror}] ] ] [$.$] ] ]
	\end{forest}}
	\caption{\label{fig:DetAdjAdv}Left: ST-Gumbel models are inconsistent in their treatment of function words and modifiers. Right: Stanford Parser trees for comparison.}
\end{figure}
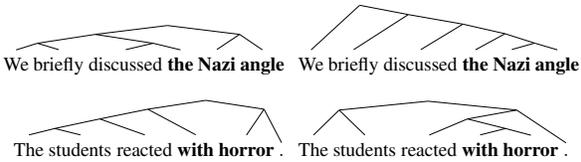

Finally, the ST-Gumbel models are not consistent in their treatment of function words, like determiners or prepositions, or in their treatment of  modifiers like adverbs and adjectives. This reflects quantitative results in Table \ref{tab:ptb} showing that ST-Gumbel parse trees correspond to PTB for PP and ADJP constituents at much lower rates than do SPINN-based models or models supplied with random trees. For example, the top left tree of Figure~\ref{fig:DetAdjAdv} (ST-Gumbel) associated the determiner \textit{the} with the verb, when it should form a constituent with the noun phrase \textit{Nazi angle} as in the top right tree (PTB). The resulting phrase, \textit{the Nazi angle}, has a clear meaning---unlike \textit{discussed the}---and it passes syntactic tests for constituency; for example, one can replace the noun phrase with the pronoun \textit{it} without otherwise modifying the meaning of the sentence. 

Similarly, prepositions are generally expected to form constituents with the noun phrases that follow them \citep{adger, sportiche2013}, as in the the bottom right tree (PTB) of Figure~\ref{fig:DetAdjAdv}. One syntactic test that \textit{with horror} forms a P-NP constituent comes from the fact that it can be a stand-alone answer to a question; for example, the question \textit{how did the students react?} can be answered simply with \textit{with horror}. ST-Gumbel models often instead pair prepositions with the verb phrases that precede them, as in Figure \ref{fig:DetAdjAdv}, lower left, where this results in the constituent \textit{the students acted with}, which cannot be a stand-alone answer to a question. From this perspective, constituents like \textit{discussed the} and \textit{we briefly} (Figure \ref{fig:DetAdjAdv}, top left) are also syntactically anomalous, and cannot be given coherent meanings.  

The ST-Gumbel models outperform syntax-based models on MultiNLI and SNLI, and the trees that they assign to sentences do not generally resemble those of PTB grammar. If we attempt to interpret these trees under the standard assumption that all the constituents in a sentence must be interpretable and must contribute to the meaning of the sentence, we force ourselves to interpret implausible constituents like \textit{we briefly}, and reach implausible sentence-level interpretations, such as taking the sentence in Figure \ref{fig:DetAdjAdv}, top left, to mean that those of us who are brief \textit{discussed the Nazi angle}. It is clear that these models do not use constituency in the same way as the widely accepted syntactic or semantic frameworks we cite do.

In sum, we find that RL-SPINN adopts a trivial, largely left-branching parse strategy, which is consistent across runs. ST-Gumbel, on the other hand, adopts the unexpected strategy to merge initial and final constituents at higher than average rates, and is also very inconsistent with its behavior on function words and modifiers. We weren't able to qualitatively identify structure that matches PTB-style syntax in ST-Gumbel parses, but we do find that it utilizes a strategy for negation---merging it with the immediately following constituent---that can lead to unexpected constituents, but nevertheless, is somewhat promising.

\section{Conclusion}

The experiments and analysis presented in this paper show that the best available models for latent tree learning learn grammars that do not correspond to the structures of formal syntax and semantics in any recognizable way. In spite of this, these models perform as well or better on sentence understanding---as measured by MultiNLI performance---as models with access to Penn Treebank-style parses. 

This result leaves us with an immediate puzzle: What do these models---especially those based on the ST-Gumbel technique---learn that allows them to do so well? We present some observations, but we are left without a fully satisfying explanation. A thorough investigation of this problem will likely require a search of new architectures for sentence encoding that borrow various behaviors from the models trained in this work.

This result also opens farther-reaching questions about grammar and sentence understanding: Will the optimal grammars for sentence understanding problems like NLI---were we to explore the full space of grammars to find them---share any recognizable similarities with the structures seen in formal work on syntax and semantics? \textit{A priori}, we should expect that they should. While it is unlikely that PTB grammar is strictly optimal for any task, the empirical motivations for many of its core constituent types---the noun phrase, the prepositional phrase, and so forth---are straightforward and compelling. However, our best latent tree learning models do not seem able to discover these structures. 

If we accept that some form of principled constituent structure is necessary or desirable, then we are left with an engineering problem: How do we identify this structure? Making progress in this direction will likely involve both improvements to the TreeRNN models at the heart of latent tree learning systems, to make sure that these models are able to perform composition effectively enough to be able to make full use of learned structures, and also improvements to the structure search methods that are used to explore possible grammars.

\section*{Acknowledgments}

This project has benefited from financial support to SB by Google, Tencent Holdings, and Samsung Research and from a Titan X Pascal GPU donated by the NVIDIA Corporation to AD. Jon Gauthier contributed to early discussions that motivated this work, and he, Nikita Nangia, Kelly Zhang, and Cipta Herwana provided help and advice. 

\bibliographystyle{acl_natbib} 
\bibliography{tacl}

\begin{thebibliography}{}
\expandafter\ifx\csname natexlab\endcsname\relax\def\natexlab#1{#1}\fi

\bibitem[{Adger(2003)}]{adger}
David Adger. 2003.
\newblock {\em Core syntax: A minimalist approach\/}.
\newblock Oxford University Press.

\bibitem[{Belinkov et~al.(2017)Belinkov, Durrani, Dalvi, Sajjad, and
  Glass}]{P17-1080}
Yonatan Belinkov, Nadir Durrani, Fahim Dalvi, Hassan Sajjad, and James Glass.
  2017.
\newblock What do neural machine translation models learn about morphology?
\newblock In {\em Proceedings of the 55th Annual Meeting of the Association for
  Computational Linguistics (Volume 1: Long Papers)\/}. Association for
  Computational Linguistics, pages 861--872.

\bibitem[{Bowman et~al.(2015)Bowman, Angeli, Potts, and
  Manning}]{bowman-EtAl:2015:EMNLP}
Samuel~R. Bowman, Gabor Angeli, Christopher Potts, and Christopher~D. Manning.
  2015.
\newblock A large annotated corpus for learning natural language inference.
\newblock In {\em Proceedings of the 2015 Conference on Empirical Methods in
  Natural Language Processing\/}. Association for Computational Linguistics,
  Lisbon, Portugal, pages 632--642.

\bibitem[{Bowman et~al.(2016)Bowman, Gauthier, Rastogi, Gupta, Manning, and
  Potts}]{bowman2016spinn}
Samuel~R. Bowman, Jon Gauthier, Abhinav Rastogi, Raghav Gupta, Christopher~D.
  Manning, and Christopher Potts. 2016.
\newblock A fast unified model for parsing and sentence understanding.
\newblock In {\em Proceedings of the 54th Annual Meeting of the {A}ssociation
  for {C}omputational {L}inguistics ({ACL})\/}. {A}ssociation for
  {C}omputational {L}inguistics, Berlin, Germany.

\bibitem[{Chen(1995)}]{chen:1995:ACL}
Stanley~F. Chen. 1995.
\newblock Bayesian grammar induction for language modeling.
\newblock In {\em Proceedings of the 33rd Annual Meeting of the Association for
  Computational Linguistics\/}. Association for Computational Linguistics,
  Cambridge, Massachusetts, USA, pages 228--235.

\bibitem[{Choi et~al.(2018)Choi, Yoo, and Lee}]{choi2017unsupervised}
Jihun Choi, Kang~Min Yoo, and Sang-goo Lee. 2018.
\newblock Learning to compose task-specific tree structures.
\newblock In {\em Proceedings of the Thirty-Second AAAI Conference on
  Artificial Intelligence (AAAI-18), to appear.\/}.

\bibitem[{Chomsky(1965)}]{chomskyaspects}
Noam Chomsky. 1965.
\newblock {\em Aspects of the Theory of Syntax\/}.
\newblock MIT press.

\bibitem[{Chung et~al.(2017)Chung, Ahn, and Bengio}]{chung2016hierarchical}
Junyoung Chung, Sungjin Ahn, and Yoshua Bengio. 2017.
\newblock Hierarchical multiscale recurrent neural networks.
\newblock In {\em Proceedings of the International Conference on Learning
  Representations ({ICLR})\/}.

\bibitem[{Cohen et~al.(2011)Cohen, Das, and Smith}]{cohen2011unsupervised}
Shay~B Cohen, Dipanjan Das, and Noah~A Smith. 2011.
\newblock Unsupervised structure prediction with non-parallel multilingual
  guidance.
\newblock In {\em Proceedings of the Conference on Empirical Methods in Natural
  Language Processing\/}. Association for Computational Linguistics, pages
  50--61.

\bibitem[{Das et~al.(1992)Das, Giles, and Sun}]{das1992learning}
Sreerupa Das, C.~Lee Giles, and Guo-Zheng Sun. 1992.
\newblock Learning context-free grammars: Capabilities and limitations of a
  recurrent neural network with an external stack memory.
\newblock In {\em Proceedings of The Fourteenth Annual Conference of Cognitive
  Science Society. Indiana University\/}. page~14.

\bibitem[{DeNero and Uszkoreit(2011)}]{denero:2011:EMNLP}
John DeNero and Jakob Uszkoreit. 2011.
\newblock Inducing sentence structure from parallel corpora for reordering.
\newblock In {\em Proceedings of the Conference on Empirical Methods in Natural
  Language Processing\/}. Association for Computational Linguistics,
  Stroudsburg, PA, USA, EMNLP '11, pages 193--203.

\bibitem[{Duda et~al.(1973)Duda, Hart, and Stork}]{duda1973pattern}
Richard~O Duda, Peter~E Hart, and David~G Stork. 1973.
\newblock {\em Pattern classification\/}.
\newblock Wiley, New York.

\bibitem[{Eriguchi et~al.(2016)Eriguchi, Hashimoto, and
  Tsuruoka}]{eriguchi-hashimoto-tsuruoka:2016:P16-1}
Akiko Eriguchi, Kazuma Hashimoto, and Yoshimasa Tsuruoka. 2016.
\newblock Tree-to-sequence attentional neural machine translation.
\newblock In {\em Proceedings of the 54th Annual Meeting of the Association for
  Computational Linguistics (Volume 1: Long Papers)\/}. Association for
  Computational Linguistics, Berlin, Germany, pages 823--833.

\bibitem[{Frege(1892)}]{fregesinn}
Gottlob Frege. 1892.
\newblock {\"U}ber sinn und bedeutung.
\newblock {\em Wittgenstein Studien\/} 1(1).

\bibitem[{Fu(1977)}]{fu77syntactic}
King~Sun Fu. 1977.
\newblock Introduction to syntactic pattern recognition.
\newblock In {\em Syntactic Pattern Recognition, Applications\/}, Springer
  Verlag, Berlin, pages 1--31.

\bibitem[{Gormley et~al.(2014)Gormley, Mitchell, Van~Durme, and
  Dredze}]{gormley-EtAl:2014:P14-1}
Matthew~R. Gormley, Margaret Mitchell, Benjamin Van~Durme, and Mark Dredze.
  2014.
\newblock Low-resource semantic role labeling.
\newblock In {\em Proceedings of the 52nd Annual Meeting of the Association for
  Computational Linguistics (Volume 1: Long Papers)\/}. Association for
  Computational Linguistics, Baltimore, Maryland, pages 1177--1187.

\bibitem[{Grefenstette et~al.(2015)Grefenstette, Hermann, Suleyman, and
  Blunsom}]{grefenstette2015learning}
Edward Grefenstette, Karl~Moritz Hermann, Mustafa Suleyman, and Phil Blunsom.
  2015.
\newblock Learning to transduce with unbounded memory.
\newblock In {\em Advances in Neural Information Processing Systems
  ({NIPS})\/}. pages 1828--1836.

\bibitem[{Heim and Kratzer(1998)}]{kratzer1998}
Irene Heim and Angelika Kratzer. 1998.
\newblock {\em Semantics in generative grammar\/}.
\newblock Blackwell.

\bibitem[{Hsu et~al.(2012)Hsu, Kakade, and Liang}]{hsu2012identifiability}
Daniel~J. Hsu, Sham~M. Kakade, and Percy~S. Liang. 2012.
\newblock Identifiability and unmixing of latent parse trees.
\newblock In {\em Advances in Neural Information Processing Systems\/}. pages
  1511--1519.

\bibitem[{Jang et~al.(2016)Jang, Gu, and Poole}]{jang2016categorical}
Eric Jang, Shixiang Gu, and Ben Poole. 2016.
\newblock Categorical reparameterization with gumbel-softmax.
\newblock In {\em Proceedings of the International Conference on Learning
  Representations ({ICLR})\/}.

\bibitem[{Joulin and Mikolov(2015)}]{joulin2015inferring}
Armand Joulin and Tomas Mikolov. 2015.
\newblock Inferring algorithmic patterns with stack-augmented recurrent nets.
\newblock In {\em Advances in Neural Information Processing Systems
  ({NIPS})\/}. pages 190--198.

\bibitem[{Kim et~al.(2017)Kim, Denton, Hoang, and Rush}]{kim2017structured}
Yoon Kim, Carl Denton, Luong Hoang, and Alexander~M. Rush. 2017.
\newblock Structured attention networks.
\newblock In {\em Proceedings of the International Conference on Learning
  Representations ({ICLR})\/}.

\bibitem[{Kingma and Ba(2015)}]{kingma2014adam}
Diederik Kingma and Jimmy Ba. 2015.
\newblock Adam: A method for stochastic optimization.
\newblock In {\em Proceedings of the International Conference on Learning
  Representations ({ICLR})\/}.

\bibitem[{Klein and Manning(2003)}]{klein2003accurate}
Dan Klein and Christopher~D. Manning. 2003.
\newblock Accurate unlexicalized parsing.
\newblock In {\em Proceedings of the 41st Annual Meeting of the {A}ssociation
  for {C}omputational {L}inguistics ({ACL})\/}. {A}ssociation for
  {C}omputational {L}inguistics, Sapporo, Japan, pages 423--430.

\bibitem[{Linzen et~al.(2016)Linzen, Dupoux, and
  Goldberg}]{linzen2016assessing}
Tal Linzen, Emmanuel Dupoux, and Yoav Goldberg. 2016.
\newblock Assessing the ability of {LSTMs} to learn syntax-sensitive
  dependencies.
\newblock {\em Transactions of the Association for Computational Linguistics\/}
  4:521--535.

\bibitem[{Liu and Lapata(2017)}]{liu2017learning}
Yang Liu and Mirella Lapata. 2017.
\newblock Learning structured text representations.
\newblock {\em Transactions of the Association for Computational Linguistics\/}
  (to appear).

\bibitem[{Maillard et~al.(2017)Maillard, Clark, and
  Yogatama}]{maillard2017jointly}
Jean Maillard, Stephen Clark, and Dani Yogatama. 2017.
\newblock Jointly learning sentence embeddings and syntax with unsupervised
  {T}ree-{LSTM}s.
\newblock {a}rXiv preprint 1705.09189.

\bibitem[{Marcus et~al.(1999)Marcus, Santorini, Marcinkiewicz, and
  Taylor}]{ptb}
Mitchell~P. Marcus, Beatrice Santorini, Mary~Ann Marcinkiewicz, and Ann Taylor.
  1999.
\newblock Treebank-3.
\newblock {LDC99T42}.
\newblock Linguistic Data Consortium.

\bibitem[{Mou et~al.(2016)Mou, Men, Li, Xu, Zhang, Yan, and Jin}]{P16-2022}
Lili Mou, Rui Men, Ge~Li, Yan Xu, Lu~Zhang, Rui Yan, and Zhi Jin. 2016.
\newblock Natural language inference by tree-based convolution and heuristic
  matching.
\newblock In {\em Proceedings of the 54th Annual Meeting of the Association for
  Computational Linguistics (Volume 2: Short Papers)\/}. Association for
  Computational Linguistics, pages 130--136.

\bibitem[{Munkhdalai and Yu(2017{\natexlab{a}})}]{munkhdalai-yu:2017:EACLlong2}
Tsendsuren Munkhdalai and Hong Yu. 2017{\natexlab{a}}.
\newblock Neural semantic encoders.
\newblock In {\em Proceedings of the 15th Conference of the European Chapter of
  the Association for Computational Linguistics: Volume 1, Long Papers\/}.
  Association for Computational Linguistics, Valencia, Spain, pages 397--407.

\bibitem[{Munkhdalai and Yu(2017{\natexlab{b}})}]{munkhdalai-yu:2017:EACLlong1}
Tsendsuren Munkhdalai and Hong Yu. 2017{\natexlab{b}}.
\newblock Neural tree indexers for text understanding.
\newblock In {\em Proceedings of the 15th Conference of the European Chapter of
  the Association for Computational Linguistics: Volume 1, Long Papers\/}.
  Association for Computational Linguistics, Valencia, Spain, pages 11--21.

\bibitem[{Naradowsky et~al.(2012)Naradowsky, Riedel, and
  Smith}]{naradowsky:2012:EMNLP}
Jason Naradowsky, Sebastian Riedel, and David~A Smith. 2012.
\newblock Improving nlp through marginalization of hidden syntactic structure.
\newblock In {\em Proceedings of the 2012 Joint Conference on Empirical Methods
  in Natural Language Processing and Computational Natural Language
  Learning\/}. Association for Computational Linguistics, pages 810--820.

\bibitem[{Naseem and Barzilay(2011)}]{naseem2011using}
Tahira Naseem and Regina Barzilay. 2011.
\newblock Using semantic cues to learn syntax.
\newblock In {\em Twenty-Fifth AAAI Conference on Artificial Intelligence\/}.

\bibitem[{Neubig et~al.(2012)Neubig, Watanabe, and Mori}]{neubig2012}
Graham Neubig, Taro Watanabe, and Shinsuke Mori. 2012.
\newblock Inducing a discriminative parser to optimize machine translation
  reordering.
\newblock In {\em Proceedings of the 2012 Joint Conference on Empirical Methods
  in Natural Language Processing and Computational Natural Language
  Learning\/}. Association for Computational Linguistics, pages 843--853.

\bibitem[{Pennington et~al.(2014)Pennington, Socher, and
  Manning}]{pennington2014glove}
Jeffrey Pennington, Richard Socher, and Christopher Manning. 2014.
\newblock Glo{V}e: Global vectors for word representation.
\newblock In {\em Proceedings of the 2014 Conference on Empirical Methods in
  Natural Language Processing ({EMNLP})\/}. {A}ssociation for {C}omputational
  {L}inguistics, Doha, Qatar, pages 1532--1543.

\bibitem[{Sag(1991)}]{sag1991}
Ivan~A. Sag. 1991.
\newblock Linguistic theory and natural language processing.
\newblock In {\em Natural language and speech: symposium proceedings,
  Brussels\/}. Springer, pages 69--83.

\bibitem[{Socher et~al.(2011)Socher, Pennington, Huang, Ng, and
  Manning}]{socher2011semi}
Richard Socher, Jeffrey Pennington, Eric~H. Huang, Andrew~Y. Ng, and
  Christopher~D. Manning. 2011.
\newblock Semi-supervised recursive autoencoders for predicting sentiment
  distributions.
\newblock In {\em Proceedings of the 2011 Conference on Empirical Methods in
  Natural Language Processing ({EMNLP})\/}. {A}ssociation for {C}omputational
  {L}inguistics, Edinburgh, UK, pages 151--161.

\bibitem[{Socher et~al.(2013)Socher, Perelygin, Wu, Chuang, Manning, Ng, and
  Potts}]{socher-EtAl:2013:EMNLP}
Richard Socher, Alex Perelygin, Jean Wu, Jason Chuang, Christopher~D. Manning,
  Andrew Ng, and Christopher Potts. 2013.
\newblock Recursive deep models for semantic compositionality over a sentiment
  treebank.
\newblock In {\em Proceedings of the 2013 Conference on Empirical Methods in
  Natural Language Processing\/}. Association for Computational Linguistics,
  Seattle, Washington, USA, pages 1631--1642.

\bibitem[{Sportiche et~al.(2013)Sportiche, Koopman, and
  Stabler}]{sportiche2013}
Dominique Sportiche, Hilda Koopman, and Edward Stabler. 2013.
\newblock {\em An introduction to syntactic analysis and theory\/}.
\newblock John Wiley \& Sons.

\bibitem[{Srivastava et~al.(2014)Srivastava, Hinton, Krizhevsky, Sutskever, and
  Salakhutdinov}]{srivastava2014dropout}
Nitish Srivastava, Geoffrey~E. Hinton, Alex Krizhevsky, Ilya Sutskever, and
  Ruslan Salakhutdinov. 2014.
\newblock Dropout: a simple way to prevent neural networks from overfitting.
\newblock {\em Journal of Machine Learning Research ({JMLR})\/}
  15(1):1929--1958.

\bibitem[{Sun et~al.(1993)Sun, Giles, Chen, and Lee}]{Sun93neural}
G.~Z. Sun, C.~L. Giles, H.~H. Chen, and Y.~C. Lee. 1993.
\newblock The neural network pushdown automaton: Model, stack and learning
  simulations.
\newblock Technical Report UMIACS-TR-93-77/CS-TR-3118, University of Maryland.

\bibitem[{Tai et~al.(2015)Tai, Socher, and Manning}]{tai2015improved}
Kai~Sheng Tai, Richard Socher, and Christopher~D. Manning. 2015.
\newblock Improved semantic representations from tree-structured long
  short-term memory networks.
\newblock In {\em Proceedings of the 53rd Annual Meeting of the {A}ssociation
  for {C}omputational {L}inguistics and the 7th International Joint Conference
  on Natural Language Processing ({ACL-IJCNLP}), Volume 1: Long Papers\/}.
  {A}ssociation for {C}omputational {L}inguistics, Beijing, China, pages
  1556--1566.

\bibitem[{Williams et~al.(2017)Williams, Nangia, and
  Bowman}]{williams2017broad}
Adina Williams, Nikita Nangia, and Samuel~R Bowman. 2017.
\newblock A broad-coverage challenge corpus for sentence understanding through
  inference.
\newblock {a}rXiv preprint 1704.05426.

\bibitem[{Williams(1992)}]{williams1992simple}
Ronald~J. Williams. 1992.
\newblock Simple statistical gradient-following algorithms for connectionist
  reinforcement learning.
\newblock {\em Machine learning\/} 8(3-4):229--256.

\bibitem[{Yogatama et~al.(2017)Yogatama, Blunsom, Dyer, Grefenstette, and
  Ling}]{yogatama2016learning}
Dani Yogatama, Phil Blunsom, Chris Dyer, Edward Grefenstette, and Wang Ling.
  2017.
\newblock Learning to compose words into sentences with reinforcement learning.
\newblock In {\em Proceedings of the International Conference on Learning
  Representations ({ICLR})\/}.

\end{thebibliography}

\end{document}